\documentclass{article} % For LaTeX2e
\usepackage{iclr2026_conference,times}

\usepackage{amsmath,amsfonts,bm}

\def\eqref#1{equation~\ref{#1}}
\def\1{\bm{1}}

\DeclareMathAlphabet{\mathsfit}{\encodingdefault}{\sfdefault}{m}{sl}
\SetMathAlphabet{\mathsfit}{bold}{\encodingdefault}{\sfdefault}{bx}{n}

\usepackage{hyperref}
\usepackage{url}
\usepackage{xurl}

\usepackage[table]{xcolor} % Required for coloring rows and cells
\usepackage{booktabs}      % For professional-looking lines (\toprule, \midrule, etc.)
\usepackage{graphicx}      % Required for \resizebox to scale the table
\usepackage{multirow}      % Required for multi-row cells in the header
\usepackage{arydshln}      % Required for dashed 
\usepackage{makecell}
\usepackage{colortbl}
\usepackage{tcolorbox}
\usepackage{fancybox}
\usepackage{enumitem}
\usepackage{tcolorbox}
\tcbuselibrary{breakable, skins}
\usepackage[table]{xcolor}
\usepackage{tikz}
\definecolor{tablegray}{gray}{0.92}
\renewcommand{\arraystretch}{1.2}
\usepackage{array}
\usepackage{hyperref}
\newcolumntype{C}[1]{>{\centering\arraybackslash}p{#1}}
\newcolumntype{R}[1]{>{\raggedleft\arraybackslash}p{#1}}
\newcolumntype{L}[1]{>{\raggedright\arraybackslash}p{#1}}
\usepackage{listings}

\title{What Shapes a Creative Machine Mind? Comprehensively Benchmarking Creativity in Foundation Models}

\author{\textbf{Zicong He}\textsuperscript{1{$\star$}},
\textbf{Boxuan Zhang}\textsuperscript{2{$\star$}},
\textbf{Weihao Liu}\textsuperscript{3{$\star$}},
\textbf{Ruixiang Tang}\textsuperscript{2},
\textbf{Lu Cheng}\textsuperscript{3}$^{\dag}$\\
$^1$ Georgia Institute of Technology\\
$^2$ Rutgers University\\
$^3$ University of Illinois Chicago\\
\texttt{zhe384@gatech.edu}\\
\texttt{\{bz362, rt836\}@rutgers.edu}\\
\texttt{\{wliu681, lucheng\}@uic.edu}\\
}

\usepackage{listings}
\iclrfinalcopy % Uncomment for camera-ready version, but NOT for submission.
\begin{document}

\maketitle

\let\svthefootnote\thefootnote
\let\thefootnote\relax\footnotetext{$^\star$ Equal contribution\hspace{3pt} \hspace{5pt}$^{\dag}$ Corresponding author\hspace{5pt}}
\let\thefootnote\svthefootnote

\begin{abstract}
The meteoric rise of foundation models (FMs) has expanded their capabilities far beyond conventional tasks. Creativity, long regarded as a hallmark of human intelligence and a driver of innovation, is now increasingly recognized as a critical dimension of machine intelligence in the era of generative FMs, complementing traditional measures of accuracy. However, existing evaluation frameworks for creativity remain fragmented, relying on ad hoc metrics that are not firmly grounded in established theories of creativity. To address this gap, we introduce $\text{C}^2$-Eval, a holistic benchmark for the unified assessment of creativity in FMs. Specifically, $\text{C}^2$-Eval distinguishes between \underline{two} complementary forms of \underline{C}reativity ($\text{C}^2$): \textbf{convergent} creativity, where tasks admit constrained solutions (e.g., code generation), and \textbf{divergent} creativity, where tasks are open-ended (e.g., story telling). It evaluates both dimensions using fine-grained assessments derived from social science theories, focusing on Usefulness, Originality, and Surprise (U-O-S). Through extensive experiments on leading proprietary and open-source models, we provide a comprehensive analysis of trade-offs in their creative capabilities. 
Our results highlight the capabilities and challenges of current FMs in pursuing a creative machine mind, showing that $\text{C}^2$-Eval provides an effective lens for examining the evolving landscape of creative AI.\footnote{Code available at \href{https://github.com/ZicongHe2002/LLM-Creativity-Benchmarking}{ https://github.com/ZicongHe2002/LLM-Creativity-Benchmarking}}
% Our results highlight critical capabilities and challenges of current FMs in how to achieve a creative machine mind, demonstrating $\text{C}^2$-Eval is an effective lens for understanding the evolving landscape of creative AI.
% both a robust benchmark for future research in machine creativity and 
% a lens for understanding the evolving landscape of creative AI.
\end{abstract}

\section{Introduction}
% Creativity, 在social science work中通常指新颖，正确，出乎意料的内容（cite some work）
% 近年来大语言模型飞速发展，在各类任务上的精度表现都十分可喜，特别是推理大模型的出现。
% The key is to frame the rise of LLMs not just as an increase in accuracy, but as an expansion of their role in the world. As they evolve from simple tools into collaborative partners, the metrics we use to judge them must also evolve. This provides a natural and compelling bridge.
% Creativity is a fundamental aspect of human intelligence, widely characterized in the social sciences as the ability to generate contents that are original, useful, and surprising (\textbf{U-O-S})

Creativity is a fundamental aspect of human intelligence, widely characterized in the social sciences as the ability to generate content that is useful, original, and surprising (\textbf{U-O-S}) \citep{simonton2012taking, amabile2018creativity}. The rapid development of Foundation Models (FMs), driven by major advances in their reasoning abilities \citep{guo2025deepseek, yang2025qwen3, openai2025o3mini}, has led to remarkable performance on a wide range of conventional tasks \citep{triviaqa_acl2017, humaneval_arxiv2021}. As FMs are increasingly positioned as collaborators in science, art, and open-ended problem-solving, only measuring their accuracy becomes fundamentally insufficient \citep{zhang2025noveltybench, creation_mmbench_arxiv2025}, but helping humans find diverse and high-quality solutions to complex problems \citep{zhao2025assessing}. Developing a rigorous framework to quantify and understand the creative potential of FMs is an essential step for responsibly guiding their development and unlocking their complete power.

Previous studies have investigated the creative potential of FMs in specific domains, such as question answering and coding \citep{he2025shakespearean, lu2024benchmarking, delorenzo2024creativeval}, as well as novel story and idea generation \citep{creation_mmbench_arxiv2025, stevenson2022gpt3aut}. While valuable, the primary limitation of these efforts is that they provide fragmented evaluations of FMs' creativity w.r.t. both the task and metric. Existing frameworks typically focus on either constrained tasks with objective answers (like code generation) or fully open-ended tasks (like story writing), overlooking the important interplay between different facets of machine creativity. Second, this fragmentation is also mirrored in how creativity itself is measured. Most evaluations assess the \textbf{U-O-S} triplet of creativity in isolation, rather than integrating them into a unified framework, failing to provide a reliable and complete understanding of the creative capabilities of modern FMs.

\begin{figure}[t]
    \centering
    \includegraphics[width=.98\textwidth]{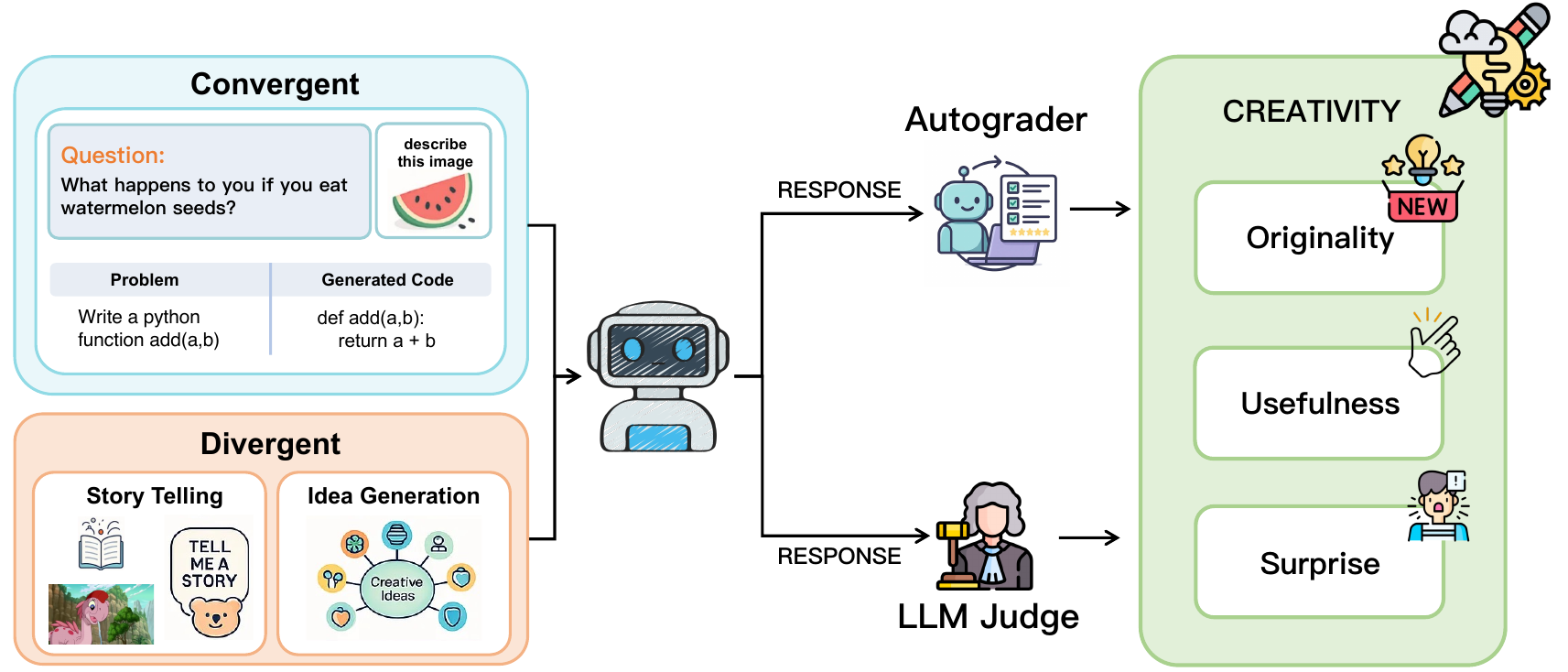}
    \caption{Overview of our $\text{C}^2$-Eval framework. \textbf{Left panel}: the two distinct creativity regimes: convergent, which focuses on structured tasks like question answering and code generation, and divergent, which evaluates open-ended tasks such as story telling and idea generation. \textbf{Right panel}: the \textbf{U-O-S} components of creativity used to evaluate responses across both regimes.}
    \label{fig:square}
    \vspace{-1em}
\end{figure}

To address these gaps, we build on the theory of creativity in social science (\textbf{U-O-S}) \citep{simonton2012taking} and introduce $\text{C}^2$-Eval, a comprehensive framework to evaluate FM creativity. We first propose a novel task-level classification of creativity: \textbf{convergent \underline{C}reativity and divergent \underline{C}reativity ($\text{C}^2$)}, as illustrated in the left panel of Figure \ref{fig:square}. Convergent creativity aims to assess an FM's ability to generate novel and varied solutions within the structured constraints of traditional tasks, such as question answering and code generation. In parallel, we define divergent creativity as the measure of the model's capacity to explore a wide range of possibilities in open-ended scenarios like story telling. Furthermore, in both regimes, every generated response is evaluated through a unified lens composed of the three core \textbf{U-O-S} components of creativity, as depicted in the right panel of Figure \ref{fig:square}. Compared to previous studies, our $\text{C}^2$-Eval framework enables a holistic and rigorous measurement of FMs creativity, capturing its innovative abilities in both structured problem solving and unconstrained generation from diverse perspectives.

Based on our $\text{C}^2$-Eval framework, we conduct an extensive evaluation of over 20 leading open-source and proprietary FMs, where we reveal several key insights. Our analysis first maps the current performance landscape, identifying the most creative proprietary and open-source models and highlighting their distinct strengths. Furthermore, our results reveal complex dynamics in model scaling; we find that larger FMs do not always have better creative performance, but that advanced reasoning capabilities appear to provide increasingly creative returns at scale. Our framework also introduces new insights about the nature of machine creativity itself by identifying a nuanced correlation between a model's convergent and divergent creativity abilities. Finally, through ablation studies, we demonstrate the significant positive influence of carefully designed creative instructions on models' output. We summarize our key contributions as follows:
\begin{itemize}
    \item We introduce $\text{C}^2$-Eval, the first comprehensive benchmark designed to systematically evaluate FMs' creativity across both \textit{convergent} and \textit{divergent} task regimes, unifying the assessment of structured tasks with open-ended generations (see Section \ref{sec:def}).
    \item We propose a principled evaluation methodology for FMs' creativity grounded in social science. It measures creativity with three classic dimensions: \textbf{ usefulness, originality, and surprise} (\textbf{U-O-S}), providing a consistent evaluation strategy (see Section \ref{sec:metrics}).
    
    \item We conduct an extensive empirical study of over 20 common closed- and open-source models, establishing a comprehensive landscape for creative AI, revealing nuanced performance trade-offs and providing a robust toolkit for guiding future research (see Section \ref{sec:results}).
\end{itemize}

% 我们首先将creativity按任务分成两类
% 一方面，我们没有忽视LLM在传统任务上优势，在事实准确性的基础上提出convergent creativity来评测LLM在既定框架下的收敛创造力。
% 另一方面，我们同样关注LLM在开放任务上的潜力，提出divergent creativity来评测LLM在开放环境下的发散创造力。
% 当待检模型生成回复后，在所有任务维度上，我们都采用了三元组评价方式来更细致，准确的衡量大模型的creativity。

\section{Related Work}
\textbf{Creativity in Social Science.}
Following classic social--psychological works, creativity is typically defined at the product level as outcomes that are both original and useful/appropriate \citep{stein1953creativity,runco2012standard}. Major reviews further emphasize that judgments of creativity are strongly contextual and socially mediated \citep{hennessey2010creativity}. 
% \citet{amabile1983social,amabile2018creativity}
 \citet{amabile1983social,amabile2018creativity}, in the componential theory, argue that domain-relevant skills, creativity-relevant processes/strategies, and intrinsic motivation (as shaped by the social environment) jointly determine creative performance; \citet{csikszentmihalyi199916,csikszentmihalyi1997flow} conceive creativity as the product of person–domain–field interactions, with the field serving as the gatekeeper of what counts as creative. We also draw on the TTCT tradition (e.g., fluency, flexibility, originality, elaboration) for backgrounded characterization \citep{torrance1966torrance,alabbasi2022educators}. Building on this consensus, we adopt \textbf{U-O-S} as our core foundational evaluation framework which remains compatible with complementary perspectives such as the Four-C model (ranging from everyday to eminent creativity) and the 4P/5A approaches (from person and process to sociocultural ecology) \citep{kaufman2009beyond,rhodes1961analysis,gluaveanu2013rewriting,boden2004creative,simonton2012taking}.

\textbf{Evaluating LLM Creativity.}
% \section{Related Work}
Existing research on LLM creativity largely follows two trajectories:  structured tasks and open-ended tasks. In structured tasks, the main practice treats functional correctness as the primary gate and adopts automated, reproducible criteria to implement two-stage protocol 'first correct, then evaluation'. In code generation, multiple samples are drawn to secure a correct answer; creativity is then assessed on the set of qualified outputs along dimensions such as fluency, flexibility, originality, and elaboration \citep{delorenzo2024creativeval}. A complementary line of work imposes incremental constraints during generation to elicit distinct solutions \citep{lu2024benchmarking}. In QA settings, once correctness is achieved, similarity metrics are used to quantify the diversity of correct solution types \citep{he2025shakespearean}.

In contrast, open-ended tasks lack a single ground truth and therefore admit more varied evaluation protocols. In the text-to-text writing-prompts setting, a prompt specifies a scenario or stylistic constraints and models generate short stories, continuations, or copy; evaluation may be performed by human raters or, increasingly, by LLM-as-Judge schemes that score coherence, style, originality, and related dimensions to improve scalability and consistency \citep{fan2018hierarchical, zheng2023judge, liu2023geval, zhao2025assessing}. In multimodal open-ended evaluation, one approach maps free-form generations to predefined options at test time \citep{liu2024mmbench}; another expands benchmarks into multi-domain, cross-disciplinary challenges to enable more robust and comprehensive assessment \citep{yue2024mmmu}. Classical psychological instruments,such as AUT and TTCT, are also employed to probe divergent creative capacity \citep{torrance2008torrance, stevenson2022gpt3aut}.
% On the structured tasks, mainstream practice treats functional correctness as the primary gate and adopts automated, reproducible criteria to implement a “first correct, then evaluation” two-stage. Multiple samples are drawn to secure correct answers, after which creativity is assessed on the set of qualified outputs along dimensions such as fluency, flexibility, originality, and elaboration \citep{10691798}. Alternatively, once correctness is achieved, similarity metrics can be used to quantify the diversity of correct solution types \citep{he2025shakespearean}. A complementary line of work imposes incremental constraints during generation to elicit distinct solutions \citep{lu2024benchmarking}.

% In contrast, open-ended tasks lack a single ground truth and therefore admit more varied evaluation protocols. One approach maps free-form generations to predefined options at test time \citep{liu2024mmbench}; another expands datasets into multi-domain, cross-disciplinary challenges to enable more robust and comprehensive assessment \citep{yue2024mmmu}. Classical psychological instruments—such as AUT and TTCT—are also employed to probe creative capacity \citep{torrance2008torrance,stevenson2022gpt3aut}. At the scoring stage, “LLM-jury” schemes have been introduced to provide more objective judgments of model outputs \citep{zhao2025assessing}.

In summary, these practices have been instrumental for probing FMs' “creativity,” yet most studies remain tailored to a single task or modality and lack an integrated, cross-task synthesis. To address this gap, we build on the \textbf{U-O-S} paradigm and propose a unified, systematized evaluation framework that standardizes judging protocols, dataset splits, and metrics across open-ended and verifiable settings, enabling comparable creativity assessments at scale.

% \begin{figure}[htbp]
%     \centering
%     \includegraphics[width=1.0\textwidth]{picture/all.pdf}
%     \caption{introduction of our whole process}
%     \label{fig:square}
% \end{figure}

\section{Methodology}

\begin{figure}[thp]
    \centering
    \includegraphics[width= 1.0\textwidth]{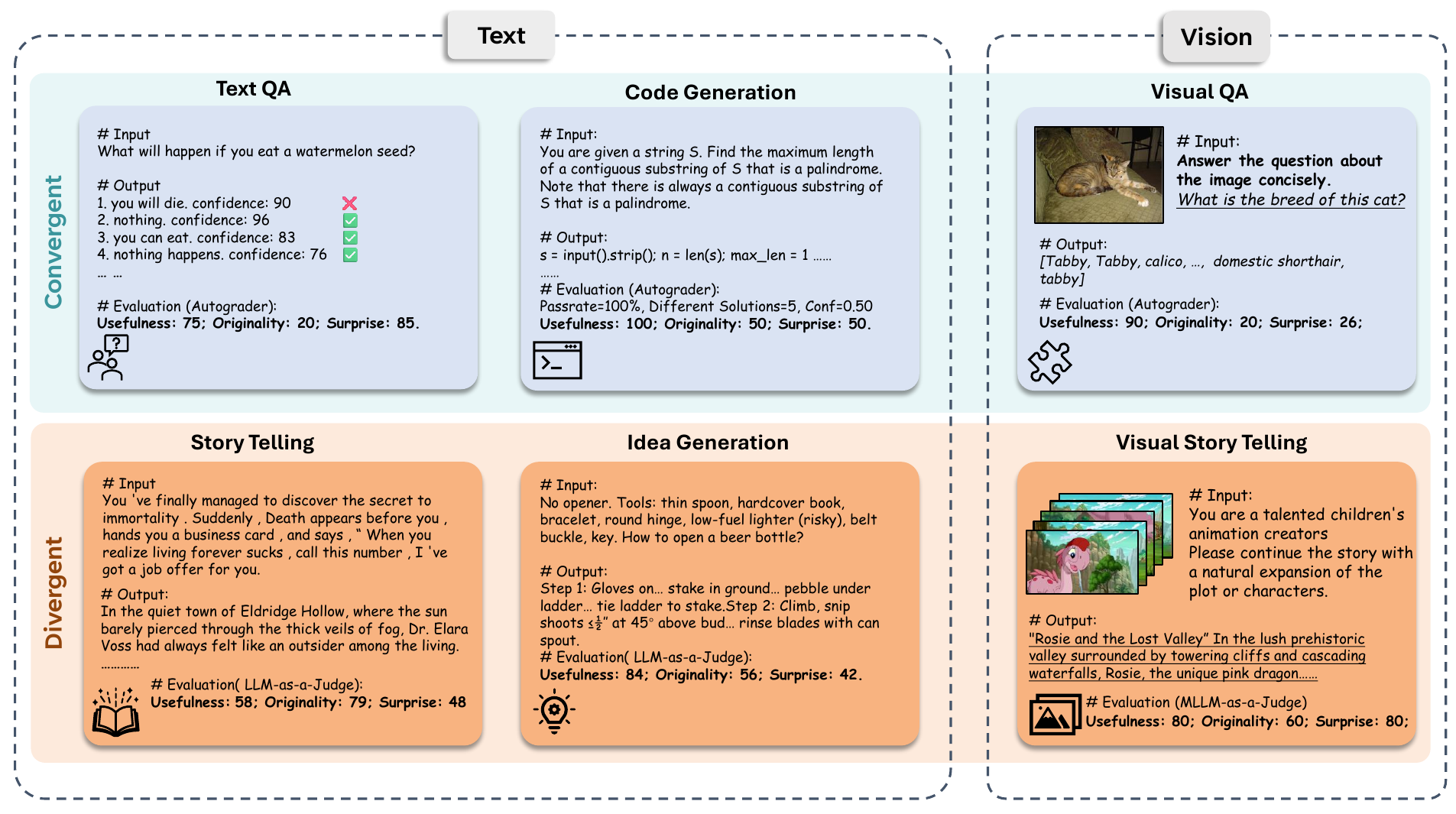}
    \caption{Illustration of Sample evaluation cases within our $\text{C}^2$-Eval framework. The \textit{top panel} presents examples of convergent creativity tasks, including Textual QA, Code Generation, and Visual QA, where an Autograder assesses the U-O-S triplet. The \textit{bottom panel} displays divergent creativity tasks, such as Story Telling, Idea Generation, and Visual Story Telling, where an advanced LLM  judges responses based on the same U-O-S components.
    Details of Autograder and LLM-as-a-Judge we used here can be found in Section ~\ref{sec:metrics}.}
    \label{fig:fig2}
    \vspace{-1em}
\end{figure}

\subsection{Definition}
\label{sec:def}
\subsubsection{The Standard Definition of Creativity}

Our framework is grounded in the standard, multi-component view of creativity established in psychological and philosophical research. The cornerstone of this view is the bipartite definition, which posits that a creative product must be both original (novel or unique) and useful (appropriate or effective for its context) \citep{runco2012standard}. Building upon this, subsequent work has incorporated a third, crucial dimension: surprise, which captures the non-obviousness or unexpectedness of an idea, distinguishing true ingenuity from straightforward problem-solving \citep{simonton2012taking}.

Aligning with these foundational treatments \citep{paul2014philosophy}, we adopt this three-component framework to define creativity. Concretely, we operationalize creativity via:

\begin{itemize}[noitemsep,nosep]
    \item \textbf{Usefulness} requires an idea to be effective and practically applicable to the task at hand;
    \item \textbf{Originality} requires an idea to be genuinely new rather than a mere rephrasing of existing solutions;
    \item \textbf{Surprise} captures a solution's non-obviousness beyond what could be achieved through conventional, algorithmic derivations.
\end{itemize}

% \subsubsection{Task Regimes: Convergent vs.\ Divergent Creativity}
\subsubsection{Task-Dependent Operationalization: Convergent vs. Divergent Creativity}

While the U-O-S triplet provides a robust conceptual foundation, its direct application across a diverse benchmark is non-trivial. The core challenge is that the method of measuring each component is fundamentally contingent on the task's objective constraints. A single, unified measurement protocol is insufficient because the very nature of creative expression differs between tasks with objective solutions versus those that are open-ended. To address this, our methodology extends beyond the standard definition by classifying tasks into two regimes, Convergent Creativity and Divergent Creativity, as shown in Figure \ref{fig:fig2}. This classification enabling a more nuanced and context-aware evaluation and dictating how we measure the three core components:

\textbf{Convergent Creativity} pertains to tasks characterized by well-defined constraints and objective correctness criteria, where the goal is to converge upon a correct solution set (e.g., factoid QA, code generation, mathematical reasoning). Within this regime, Usefulness is mapped to correctness, Originality to the diversity among correct solutions, and Surprise to the non-obviousness of a given correct answer. Here, creativity often manifests in the problem-solving pathway.

\textbf{Divergent Creativity} addresses open-ended tasks lacking a single ground-truth answer, encouraging the exploration of a vast solution space (e.g., unimodal and multimodal story continuation). In this context, where objective correctness is absent, Usefulness is interpreted as appropriateness and coherence, Originality as perceived novelty, and Surprise as unexpectedness. Here, creativity is embodied directly within the generated content itself.

\subsection{Evaluation Metrics}
\label{sec:metrics}

\subsubsection{Convergent Creativity}
\label{autograder}

The following pipeline illustrates our autograder. For each prompt in the convergent creativity regime, we generate a set of $n$ candidate outputs, denoted by $\{y_i\}_{i=1}^n$. To evaluate them, we first introduce an indicator function, $\mathbb{I}(y_i) \in \{0, 1\}$, which returns 1 if the output $y_i$ is correct according to the task's specific constraints, and 0 otherwise.

\textbf{Usefulness.}
In the convergent regime, usefulness is synonymous with correctness. We therefore define the \textit{usefulness} score, $U$, as the accuracy over the $n$ generated samples: $U = \frac{1}{n} \sum_{i=1}^{n} \mathbb{I}(y_i)$.
% = \frac{c}{n},$ where $c$ is the total count of correct outputs.

\textbf{Originality.}
We measure the originality of the generated response as the semantic diversity among the set of correct solutions. To this end, we partition the correct outputs into disjoint clusters using a threshold-based rule: any two correct outputs $y_i$ and $y_j$ are assigned to the same cluster if their cosine similarity meets or exceeds a pre-defined threshold $\tau$ (e.g., 0.9). Each problem $p$ yields a count $K_p$ of distinct clusters, and the prompt-level \textit{originality} score is defined as $O= K_p$.

\textbf{Surprise.}
We define Surprise as the complement of the model's own confidence-of-correctness estimate: lower self-reported confidence implies higher Surprise. 
Given a prompt $x$, we generate $n$ candidate answers $\{y_i\}_{i=1}^{n}$ via multi-sample sampling. 
After producing each $y_i$, we then instruct the model to generate a confidence score $c_i \in [0,1]$ that indicates the probability that $y_i$ is correct. \citep{xiong2023can} Let $z_i \in \{0,1\}$ denote the ground-truth correctness of $y_i$. 
Following \citet{kadavath2022language}, the prompt-level confidence averages confidence over correct samples only by
\[
C = \frac{\sum_{i=1}^{n} z_i\, c_i}{\max\left(1,\; \sum_{i=1}^{n} z_i\right)}.
\]
We then define the prompt-level \textit{surprise} metric as the complement:
$S \;=\; 1 - C.$

% \subsection{Divergent Creativity}
% In open-ended generation without a single ground truth (e.g., story continuation), we use an automatic rater (e.g., GPT-5) to produce three scalar scores in \([0,1]\) per sample aligned with our definitions:
% \(\,U_{\text{fit}}\) (appropriateness/fit), \(N_{\text{orig}}\) (perceived originality), and \(S_{\text{surp}}\) (unexpectedness).
% Optionally, to reflect sample-to-sample variety, we also cluster the \(K\) samples for the same prompt using the cosine-similarity rule above and compute the distinct-cluster ratio \(N_{\text{clus}}=K/\max(1,K_{\text{all}})\); when used, the final Originality is the mean of the two:
% % \begin{equation}
% % N \;=\; \frac{N_{\text{orig}} + N_{\text{clus}}}{2}, \qquad
% % U \;=\; U_{\text{fit}}, \qquad
% % S \;=\; S_{\text{surp}}.
% % \end{equation}

\subsubsection{Divergent Creativity}

% For open-ended generation tasks that lack a single ground-truth answer , we employ an automatic rater (GPT-5 \& GPT-4o) to evaluate each generated sample\citep{fan2018hierarchical, zheng2023judge, liu2023geval, zhao2025assessing}. This rater produces three scalar scores in $[0, 1]$, aligned with our core definitions.

% The \textbf{Usefulness ($U$)} and \textbf{Surprise ($S$)} scores for a given prompt are calculated as the average of the rater's scores across all $K_{\text{total}}$ generated samples:
% \begin{itemize}
%     \item \textbf{Usefulness}: The average perceived appropriateness and coherence, denoted as $U_{\text{usefulness}}$.
%     \item \textbf{Surprise}: The average perceived unexpectedness, denoted as $S_{\text{surprise}}$.
%     \item \textbf{Originality.} Average perceived degree to which the work presents novel, creative, or unexpected ideas that go beyond clichés—evaluated relative to the prompt; denoted $O_{\text{originality}}$.
% \end{itemize}

For open-ended generation tasks that do not have a single ground-truth answer, we employ an automatic rater, generally the most advanced judge model \citep{tan2024judgebench}, to score each generated sample \citep{fan2018hierarchical, zheng2023judge, liu2023geval, zhao2025assessing}. To ensure a standardized and rigorous evaluation, the rater is provided with a detailed prompt containing explicit definitions and multi-point scoring rubrics for each of our three core creativity components.

The evaluation proceeds in two stages. First, for a given input prompt, we generate $K_{\text{total}}$ distinct output samples. Each of these samples is then individually evaluated by the automatic rater, yielding a sample-level score triplet $(U_i, O_i, S_i) \in [0, 5]^3$ for each of the $i=1, \dots, n$ samples. These scores (the larger the better) reflect the following conceptual underpinnings:

\begin{itemize}[noitemsep,nosep]
    \item \textbf{Usefulness ($U$)} measures the coherence, logical consistency, and contextual appropriateness of the generated output. It assesses whether the response is a well-structured and meaningful continuation of the prompt, free of contradictions or nonsensical elements.

    \item \textbf{Originality ($O$)} assesses the novelty of the ideas presented. It quantifies the degree to which the output moves beyond clichés, predictable narrative paths, and conventional tropes to introduce fresh, imaginative concepts.

    \item \textbf{Surprise ($S$)} captures the non-obviousness and emotional engagement of the output. It measures the presence of unexpected plot twists, striking imagery, or insightful developments that make the generation compelling and memorable.
\end{itemize}

\setlength{\tabcolsep}{3pt}
\begin{table}[t]
\centering
\scriptsize
\renewcommand{\arraystretch}{1.2} % 增加行间距
\resizebox{0.99\linewidth}{!}{\begin{tabular}{C{0.125\textwidth} !{\color{gray}\vrule width 0.5pt} c !{\color{gray}\vrule width 0.5pt} 
cccc !{\color{gray}\vrule width 0.5pt} cccc !{\color{gray}\vrule width 0.5pt} cccc !{\color{gray}\vrule width 0.5pt} cccc}
\toprule
\multirow{3}{*}{Model Name} & \multirow{3}{*}{\makecell{Combined\\Creativity}} 
& \multicolumn{8}{c !{\color{gray}\vrule width 0.5pt}}{Convergent Creativity} 
& \multicolumn{8}{c}{Divergent Creativity} \\
\cline{3-18}
 & & \multicolumn{4}{c !{\color{gray}\vrule width 0.5pt}}{Question Answering} & \multicolumn{4}{c !{\color{gray}\vrule width 0.5pt}}{Code Generation} 
 & \multicolumn{4}{c !{\color{gray}\vrule width 0.5pt}}{Story Telling} & \multicolumn{4}{c}{Idea Generation} \\
\cline{3-18}
 & & Use. & Orig. & Surp. & Crea. & Use. & Orig. & Surp. & Crea. 
 & Use. & Orig. & Surp. & Crea. & Use. & Orig. & Surp. & Crea. \\
\Xhline{1pt}
Qwen2.5-7B                                         & 31.9                                                                                           & 52.8       & \underline{13.9}        & 19.5       & 28.7       & 22.7      & 30.0       & 25.5    & 26.1      & 52.7      & 53.4      & 39.9      & 48.7      & 25.3     & 22.3       & 15.0       & 23.9      \\
Qwen2.5-14B                                        & 35.0                                                                                           & 67.6       & 13.1        &11.8       & 30.8       & 43.9   & 36.8    & \textbf{31.0}    & 37.2   & 55.1   & 61.9   & 40.5   & 52.5   & 21.1      & 17.1       & 11.8       & 19.5      \\
Qwen2.5-32B                                        & 35.4                                                                                           & 66.7       & 6.2        & 16.2       & 29.7       & 48.5   & 49.4    & 14.7    & 37.5   & 48.6   & 63.4   & 34.3   & 48.8   & 26.4    & 24.5       & 17.5      & 25.7      \\
Qwen2.5-72B                                        & 35.3                                                                                           & 75.5       & 7.9        &  6.7      & 30.0       & 56.6   & 43.4    & 15.2    & 38.4   & 44.7   & 66.2   & 31.6   & 47.5  & 27.2      & 21.4       & 15.3       & 25.1      \\
*Qwen3-8B                                           & 40.2                                                                                           & 59.7       & 8.4        & 17.5       & 28.5       & 84.2   & 70.2    & 20.8    & 58.4   & 66.8      & 71.0      & 57.4      & 65.1      & 29.4      & 31.1       & 21.2       & 29.4      \\
*Qwen3-235B                                         & \textbf{58.3}                                                                                           & 86.6       & 12.2        &  9.6      & 36.1       & \textbf{91.2}      & \textbf{78.4}       & 22.8       & \textbf{64.1}      & \textbf{85.5}      & \textbf{90.7}      & \textbf{77.6}      & \textbf{84.6}      & 52.4      & \underline{42.8}     & \underline{33.0}       & 48.4      \\
*QwQ-32B                                            & 54.3                                                                                           & 70.4       & \textbf{21.8}       & \underline{19.7}       & \underline{37.3}       & 77.9   & 62.9    & \underline{28.3}    & 56.4   & 78.7   & 83.3   & \underline{71.2}   & 77.9   & 48.7      & 40.7       & 29.9       & 45.5      \\
Gemma3-27B                                         & 40.7                                                                                           & 76.0      & 10.7        & 5.6       & 30.8       & 61.1      & 24.7       & 6.6       & 30.8      & 75.0      & 83.0      & 67.5      & 75.2      & 28.7      & 21.9    & 15.0       & 26.1      \\
*DeepSeek R1                                        & 54.5                                                                                           & \underline{87.4}       & 5.0        &  6.3      & 32.9       & 79.3      & 66.4       & 20.7       & 55.5      & \underline{80.3}      & \underline{87.5}      & 70.8      & \underline{79.5}      & \underline{54.5}  & 42.1    & 32.3       & \underline{49.9}      \\
\rowcolor{gray!20} Claude3.7      & 50.3                                                                                           & 84.6       & 10.6        & 16.7       & \underline{37.3}       & 78.0   & 54.7    & 16.9    & 49.9   & 67.9   & 85.8   & 59.5   & 71.1   & 46.4      & 37.4       & 27.8       & 43.1      \\
\rowcolor{gray!20}*o4-mini & \underline{57.7}                                                                                           & 86.6       & 8.1        & 17.1       &  \textbf{37.3}      & \underline{86.2}   & \underline{70.9}    & 19.5    & \underline{58.9}   & 78.1   & 86.6   & 66.8   & 77.2   & \textbf{57.0}      & \textbf{57.4}       & \textbf{46.5}       & \textbf{57.4}      \\
\rowcolor{gray!20} GPT-4o         & 43.0                                                                                           & \textbf{89.0}       & 9.6        & 8.8       & 35.8       & 66.8   & 54.6    & 20.7    & 47.4   & 54.8   & 73.4   & 37.9   & 55.4   & 37.9      & 26.1       & 18.6       & 33.5      \\
\rowcolor{gray!20} GPT-4o-mini    & 37.0                                                                                           & 73.5       & 8.7        & \textbf{19.8}       & 34.0       & 48.4   & 39.1    & 13.6    & 33.7   & 53.2   & 69.1   & 39.0   & 53.8   & 29.7      & 21.3       & 14.2       & 26.6      \\
\bottomrule
\end{tabular}}
% \caption{Metrics: QA(LLM-as-a-Judge), Code Gen(AutoGrader), Story Telling(LLM-as-a-Juedge), Idea Generation(LLM-as-a-Judge)}
\vspace{-2mm}
\caption{Evaluation results of mainstream open-source and proprietary models on text-based tasks. Adopted metrics are usefulness~(Use.), originality~(Orig.), surprise~(Surp.), and creativity~(Crea.). Combined Creativity is the average creativity score of models on both convergent and divergent tasks, showing a model's comprehensive creativity. All scores are in $[0,100]$. \textbf{Bold} numbers are models with the best corresponding performance, and the \underline{underlined} ones are the corresponding second-best. (*) represents a reasoning model with thinking process.}
\vspace{-2mm}
\label{tab:main_text}
\end{table}

\begin{table}[h]
\centering
\scriptsize
\renewcommand{\arraystretch}{1.2} % 增加行间距
{\begin{tabular}{c !{\color{gray}\vrule width 0.5pt} c !{\color{gray}\vrule width 0.5pt} 
C{0.08\textwidth}C{0.08\textwidth}C{0.08\textwidth}C{0.08\textwidth} !{\color{gray}\vrule width 0.5pt} C{0.08\textwidth}C{0.08\textwidth}C{0.08\textwidth}C{0.08\textwidth}}
\toprule
\multirow{3}{*}{Model Name} & \multirow{3}{*}{\makecell{Combined\\Creativity}} 
& \multicolumn{4}{c !{\color{gray}\vrule width 0.5pt}}{Convergent Creativity} 
& \multicolumn{4}{c}{Divergent Creativity} \\
\cline{3-10}
 & & \multicolumn{4}{c !{\color{gray}\vrule width 0.5pt}}{Vision Question Answering} & \multicolumn{4}{c}{Visual Story Telling} \\
\cline{3-10}
 & & Use. & Orig. & Surp. & Crea. & Use. & Orig. & Surp. & Crea. \\
\Xhline{1pt}
Qwen2.5-VL-32B & 48.4 & 58.8 & 15.4 & 12.6 & 28.9 & 63.8 & \underline{85.0} & 54.6 & 67.8 \\
Deepseek-VL2& 46.3 & 66.3 & 16.6 & 18.8 & 33.9 & 53.4 & 79.4 & 43.0 & 58.6 \\
Qwen2.5-VL-72B & 42.0 & 54.4 & 14.1 & 10.2 & 26.2 & 48.6 & 79.2 & 45.4 & 57.8 \\
Intern3-VL-78B & 38.8 & 29.4 & 15.1 & 18.6 & 21.0 & 48.8 & 79.8 & 41.0 & 56.6 \\
Llama3.2-11B-V & 34.6 & 25.9 & 12.1 & 4.1 & 14.0 & 49.6 & 73.8 & 42.4 & 55.2 \\
Qwen2.5-VL-7B & 38.7 & 42.4 & 10.7 & \underline{19.5} & 24.2 & 46.6 & 74.4 & 38.4 & 53.2 \\
\rowcolor{gray!20} Claude3.7  & \underline{56.6} & \underline{70.0} & \underline{28.0} & 12.6 & \underline{36.9} & \textbf{70.2} & \textbf{93.0} & \textbf{65.5} & \textbf{76.2}  \\
\rowcolor{gray!20} o4-mini & \textbf{59.5} & \textbf{73.0} & \textbf{30.6} & \textbf{36.2} & \textbf{46.6} & \underline{67.2} & 90.6 & \underline{59.4} & \underline{72.4}  \\
\rowcolor{gray!20} Claude3.5 & 46.9 & 62.5 & 19.4 & 7.9 & 29.9 & 61.6 & 82.6 & 47.4 & 63.8  \\
\rowcolor{gray!20} GPT-4o-mini & 43.9 & 51.5 & 14.3 & 10.2 & 25.3 & 55.0 & 81.8 & 50.6 & 62.4 \\
\bottomrule
\end{tabular}}
\vspace{-1mm}
\caption{Experimental results of VLMs on multimodal tasks. All scores are in $[0,100]$. \textbf{Bold} and \underline{underlined} fonts indicate the best and second-best performing models.}
\label{tab:main_vision}
\vspace{-2em}
\end{table}

Finally, we aggregate these sample-level scores to produce a stable, \textbf{prompt-level} score for each metric by taking the average across all samples, as described by the unified aggregation formula: $M_{\text{prompt}} = \frac{1}{n} \sum_{i=1}^{n} M_i, \quad \text{where } M \in \{\mathrm{U}, \mathrm{O}, \mathrm{S}\}.$ 
% This process both captures the quality of individual generations at the sample level and provides an aggregated measure of a model's creative performance for a given prompt.
    
\textbf{Overall Creativity Score.} For convergent and divergent creativity, we first scale the range of evaluated values into [0, 100], obtaining a triplet $(\mathrm{U}, \mathrm{O}, \mathrm{S}) \in [0, 100]^3$. The score range is selected for convenient comparison and interpretation. We then compute the final composite creativity score, $C$, by taking the average of these three components.
\section{Benchmarking Creativity in FMs}
\subsection{Setup}
\label{sec:setup}

\paragraph{Models.} 

% open-source models
% proprietary models
% \citep{lu2024deepseek}
% We evaluate five series of models: \textbf{Qwen} (Qwen2.5 7B--72B; Qwen3-8B; QwQ-32B~\citep{qwen25_blog,qwq32b_blog}), \textbf{GPT} (GPT-4o, GPT-4o-mini, o4-mini~\citep{openai_gpt4omini_blog,openai_gpt4o_syscard}), \textbf{Claude} (Claude~3.5/3.7 Sonnet~\citep{claude35_blog,claude37_blog}), \textbf{DeepSeek} (DeepSeek-R1~\citep{guo2025deepseek}), and \textbf{Gemma} (Gemma 3-27B~\citep{team2025gemma}).
To provide a comprehensive assessment of creative capabilities, we evaluate a diverse cohort of both proprietary and open-source models across text-only and vision-language paradigms.

Our evaluation of proprietary models encompasses recent releases from OpenAI and Anthropic. From OpenAI, we include GPT-4o, GPT-4o-mini, and o4-mini model and both of GPT-4o-mini and o4-mini models are evaluated in text-only and vision-language contexts~\citep{openai_gpt4omini_blog,openai_gpt4o_syscard}. Similarly, from Anthropic, we test Claude 3.5 Sonnet and Claude 3.7 Sonnet, both of which are also evaluated across both modalities to gauge their multimodal creative performance~\citep{claude35_blog,claude37_blog}.

In the open-source domain, our selection is categorized by modality. For text-only tasks, we include Qwen2.5 (7B--72B; Instruct), Qwen3-8B (thinking and non-thinking), QwQ-32B~\citep{qwen25_blog,qwq32b_blog}, DeepSeek-R1~\citep{guo2025deepseek}, and Gemma 3-27B~\citep{team2025gemma}. Our open-source vision-language models (VLMs) consist of Qwen2.5-VL (7B, 32B, 72B; Instruct)~\citep{qwen25vl_github}, DeepSeek-VL2~\citep{lu2024deepseek}, InternVL3-78B~\citep{zhu2025internvl3}, and Llama-3.2-11B-Vision~\citep{meta2024llama}.

% \paragraph{Models.}
% We evaluate both open-source and proprietary models across text-only and vision–language settings to assess creativity under varied capacities and modalities. For text-only open-source models, we include Qwen2.5 (7B--72B; Instruct), Qwen3-8B (thinking and non-thinking), QwQ-32B~\citep{qwen25_blog,qwq32b_blog}, DeepSeek-R1~\citep{guo2025deepseek}, and Gemma 3-27B~\citep{team2025gemma}. For text-only proprietary models, we evaluate GPT (GPT-4o, GPT-4o-mini, o4-mini)~\citep{openai_gpt4omini_blog,openai_gpt4o_syscard} and Claude (Claude 3.5 and 3.7 Sonnet)~\citep{claude35_blog,claude37_blog}.

% For vision–language open-source models, we adopt Qwen2.5-VL (7B, 32B, 72B; Instruct)~\citep{qwen25vl_github}, DeepSeek-VL2~\citep{lu2024deepseek}, InternVL3-78B~\citep{zhu2025internvl3}, and Llama-3.2-11B-Vision~\citep{meta2024llama}. For proprietary VLMs, we include Claude (Claude 3.5 and 3.7 Sonnet)~\citep{claude35_blog,claude37_blog} and GPT (o4-mini and GPT-4o-mini)~\citep{openai_gpt4omini_blog,openai_gpt4o_syscard}.

\textbf{Datasets.}
Convergent tasks admit clear, verifiable answers or functional goals, whereas divergent tasks do not assume a single ground truth and emphasize novelty and appropriateness. For convergent creativity, we evaluate three kinds of tasks with ground truth: \emph{Question Answering (QA)}, \emph{Code Generation}, and \emph{Visual Question Answering (VQA)} . For QA, we use \textbf{TriviaQA} \citep{triviaqa_acl2017}, which is a large-scale, evidence-grounded reading comprehension and open-domain QA from real queries. For Code Generation,  we use \textbf{LiveCodeBench} ~\citep{livecodebench_openreview2024} to provide program synthesis benchmarks with unit tests that enable objective and reproducible correctness checks. For VQA, we use \textbf{OK-VQA} ~\citep{marino2019ok}, which pairs near-duplicate images to reduce language priors and foreground genuine visual understanding in QA. 
For divergent creativity, we evaluate on two different open-ended generation tasks, Story Telling and Idea Generation. 
For textual story telling, we utilize \textbf{WritingPrompts}~\citep{writingprompts_hf}, which consists of prompt–story pairs for free-form narrative generation.
For visual story telling,  we adopt the stories introduced in \textbf{Creation-MMBench}~\citep{creation_mmbench_arxiv2025} to assess image-grounded creative generation.  
We evaluate the idea generation capability on \textbf{MacGyver}~\citep{macgyver_acl}, which presents constrained, real-world–inspired scenarios that require multi-step reasoning and inventive use of resources to achieve feasible solutions.
% \emph{Visual Story Telling} with \textit{Creation-MMBench}, \emph{Story Telling} with \textit{WritingPrompts}, and \emph{Idea Generation} with \textit{MacGyver}. 

\subsection{Results and Discussions}
\label{sec:results}
% \subsubsection{Convergent Creativity in LLMs}

\begin{figure}[t]
    \centering
    \includegraphics[width= 1.0\textwidth]{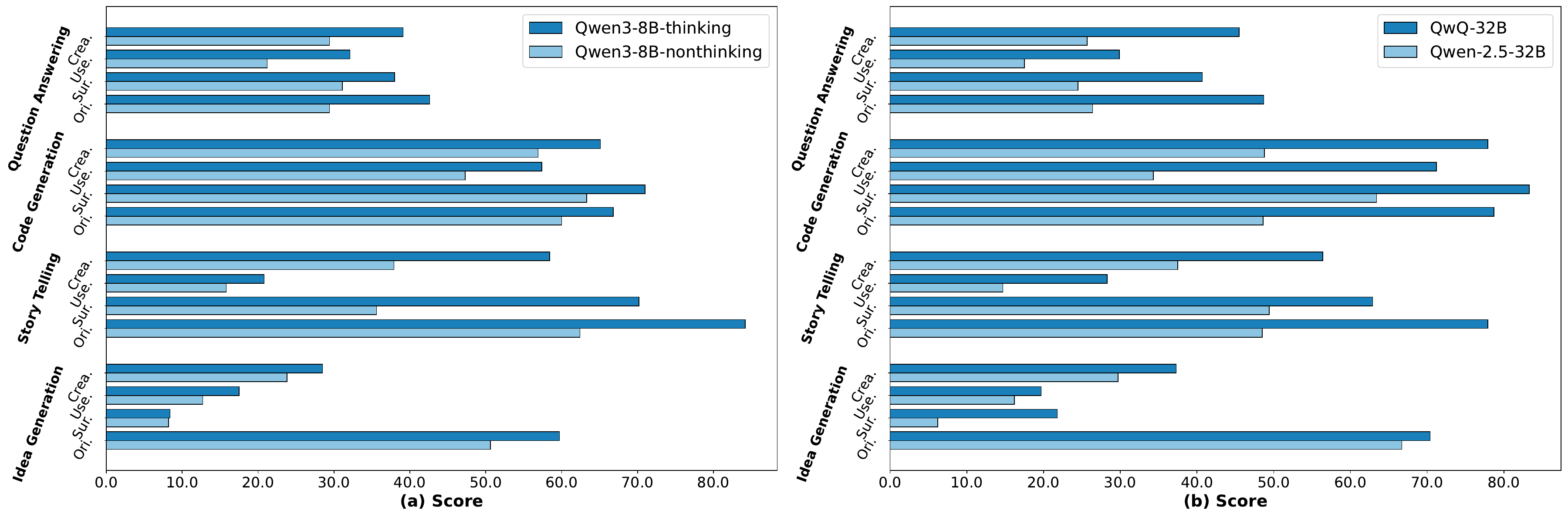}
    \caption{Reasoning vs. non-reasoning models of the same size. \textbf{Left (a):} Qwen3-8B in thinking mode versus Qwen3-8B in non-thinking mode. \textbf{Right (b):} Qwen2.5-32B vs QwQ32B.}
    \label{fig:compare}
\end{figure}

\subsubsection{Main Results}

% Vision 和 Text 的共同特征
% \paragraph{}
% \paragraph{Effectiveness of Chain-of-Thought Reasoning (CoT).}

% Within the same model family (see Table~1/2), we compare standard and CoT inference styles and observe: (i) the number of correct solutions (i.e., Usefulness) and Originality steadily increase, as reflected by the marked rise in Originality; (ii) within matched accuracy bins, Surprise remains at a comparable level without a systematic drop. These results indicate that explicit reasoning expands paths and solution space without sacrificing non-obviousness, thereby enhancing novelty and usefulness while maintaining appropriateness.
\paragraph{Which FM has the most creative mind?} For text-based tasks, we can see from Table~\ref{tab:main_text} that \textit{Qwen3-235B} and \textit{o4-mini} are the most and the second most creative models. Qwen3-235B achieves the best performance on code generation and story telling, showing its expertise in long-tail creative content generation. While in QA and idea generation tasks, whose outputs are typically shorter, o4-mini performs better than other models. These results indicate that open-source LLMs can hold comparable and even better creativity than commercial LLMs.
% Tasks 相关的, convergent&divergent 相关的
For vision-based tasks, our results summarized in Table \ref{tab:main_vision} reveal that proprietary models currently possess the most creative capabilities. \textit{o4-mini} achieves the highest combined creativity score (59.5), illustrating its remarkable talent for ingenuity within the structured constraints of the VQA task. In parallel, \textit{Claude 3.7} demonstrates superior imaginative fluency, dominating the open-ended Visual Story Telling task. A clear creativity gap separates these leading proprietary models from the top open-source models. Among the open-source ones, \textit{Qwen2.5-VL-32B} is the top performer due to its strong narrative generation, though it falls short in producing diverse solutions for constrained problems. The second-best open-source model, \textit{Deepseek-VL2}, displays the opposite strength profile, excelling in convergent creativity while being less skilled in divergent story telling.

\textbf{Larger sizes do not always mean greater creativity.}
% Creativity is not monotonic in size. On the \(U\!-\!O\!-\!S\) panel, larger models are not necessarily more creative. In the LLLM-based evaluation, the \textit{Qwen2.5-32B} and \textit{Gemma3-27B} outperform larger models (\textit{Qwen2.5-72B}). This reflects a normative convergence effect: once the correctness threshold is crossed, larger (especially more preference-aligned) models tend to produce more concentrated, standardized, and conservative solutions, compressing \textit{Originality} and \textit{Surprise}.
Creativity does not increase monotonically with model size. In the U-O-S panel, larger models are not necessarily more creative. In the LLM-based evaluation, within the same model family, \textit{Qwen2.5-32B} attains a combined creativity score of 35.4, slightly higher than the larger \textit{Qwen2.5-72B} at 35.3; likewise, \textit{Gemma3-27B} at 40.7 clearly exceeds the score of \textit{Qwen2.5-72B}. In the VLM-based evaluation, this phenomenon is most clear within the open-source \textit{Qwen2.5-VL} family, where the 32B variant (48.4) significantly outperforms its much larger 72B counterpart (42.0) in combined creativity. Similarly, the \textit{Intern3-VL-78B} model is surpassed by the smaller 32B Qwen model. This suggests there may be an optimal balance in model size for fostering creativity, and simply increasing the parameter count is not a guaranteed path to a more creative machine mind.

\textbf{Reasoning yields larger marginal gains in creativity at scale.} Reasoning models often present impressive performances in creativity evaluation. For example,\textit{Qwen3-8B-thinking}, \textit{Qwen3-235B}, \textit{QwQ-32B}, \textit{DeepSeek R1} and \textit{o4-mini} always outperform in creativity scores. Notably, scaling up the length of output tokens appears more important than scaling up the size of the model w.r.t. increasing FMs' creativity. As we can see in Figure \ref{fig:compare}, reasoning models (QwQ-32B,Qwen3-8B-thinking) significantly outperform non-reasoning models (Qwen32B,Qwen3-8B-nonthinking) under all creativity metrics, though they have the same model size.

\begin{figure}[t]
    \centering
    \includegraphics[width= 1.0\textwidth]{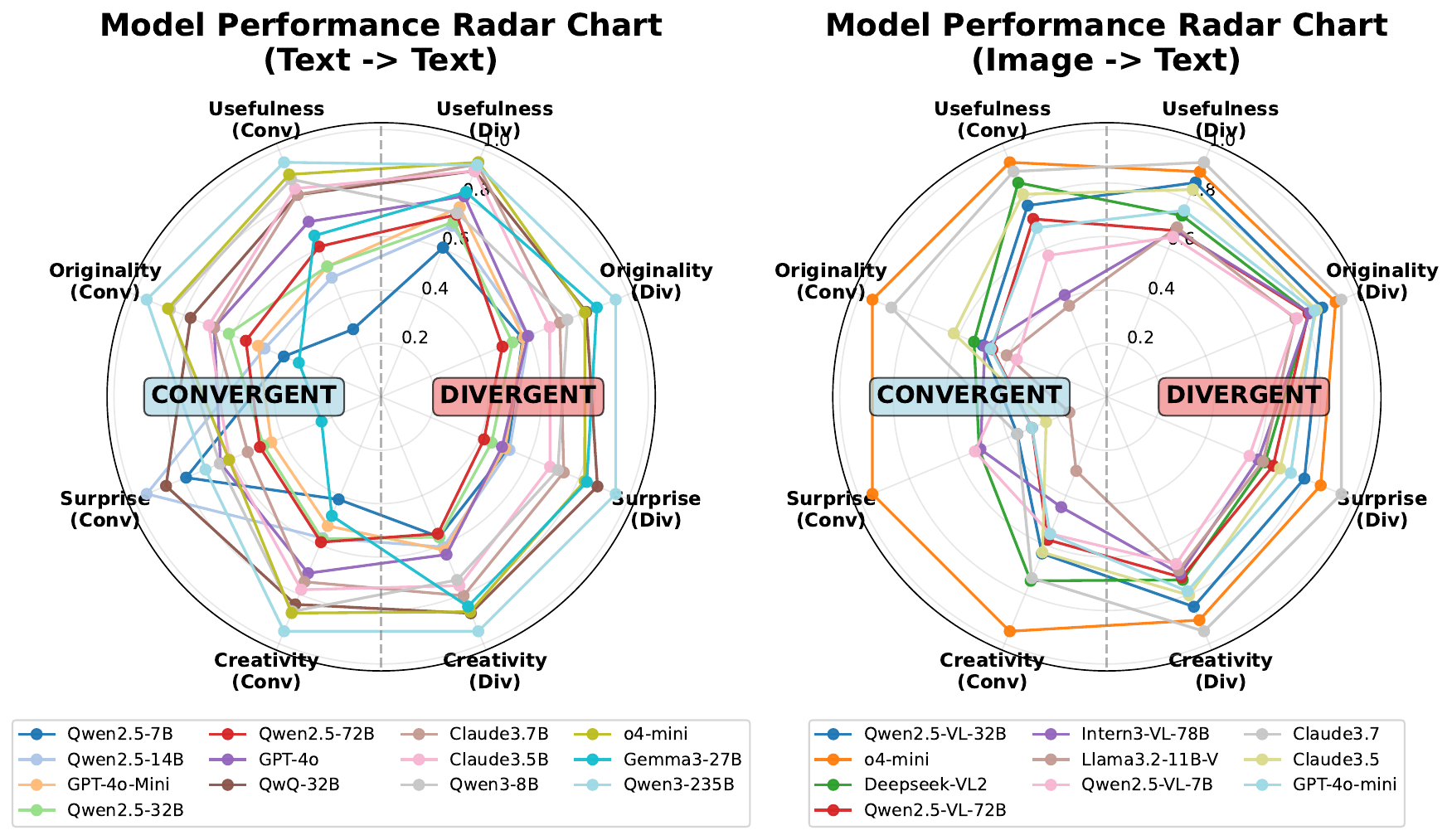}
    \caption{Radar chart illustrates the multi-dimensional components of creativity. The eight axes in each sub-figure represent four indicators (usefulness, originality, surprise, creativity) under both divergent (Div.) and convergent creativity (Conv.). All values here are normalized to (0,1). }
    \label{fig:radar}
\end{figure}

% \begin{figure}[thp]
%     \centering
%     \includegraphics[width= 0.98\textwidth]{picture/temperature_analysis.pdf}
%     \caption{TBD}
%     \label{fig:temper}
% \end{figure}

% \paragraph{}

\subsubsection{Correlation between Divergent and Convergent Creativity}
To understand whether the capabilities that drive creative performance in structured problem-solving also translate to open-ended generation, we investigate the correlation between convergent and divergent creativity across different models and modalities and present two radar charts in Figure \ref{fig:radar}. For the text generation tasks (left radar chart), our analysis encompasses convergent tasks such as Question Answering (using TriviaQA) and Code Generation (using LiveCodeBench), as well as divergent tasks like Textual Story Telling (using WritingPrompts) and Idea Generation (using MacGyver). For the visual language tasks (right radar chart), we analyze convergent Visual Question Answering (on OK-VQA) and divergent Visual Story Telling (on Creation-MMBench).

\paragraph{Varied U-O-S Strengths Across Textual Tasks.} For the text to text generation tasks, the left radar chart of Figure \ref{fig:radar} reveals that models often display varied strengths across the U-O-S dimensions when moving between convergent and divergent tasks. It illustrates that a model might exhibit high usefulness in structured, convergent tasks like code generation, but a greater capacity for originality in open-ended, divergent tasks such as story telling. This suggests that the mechanisms supporting creativity are not always uniformly leveraged across different task types, leading to distinct creative profiles and trade-offs in performance.

\paragraph{Diverse Multimodal Creativity Profiles.} Similarly, the right radar chart of Figure \ref{fig:radar}, focusing on the image to text generation tasks, demonstrates a complex relationship between convergent and divergent creativity in multimodal contexts. Models show diverse performance distributions across the U-O-S components for tasks like visual question answering (convergent) and visual story telling (divergent). For instance, while \textit{o4-mini} exhibits strong U-O-S performance in both regimes, \textit{Claude3.7} demonstrates a particularly pronounced increase in U-O-S when generating visual stories, suggesting a stronger capacity for imaginative fluency in divergent creativity. This highlights that the balance between these creative attributes often varies depending on the specific task demands.

\subsubsection{Influence of Creative Instructions}

To understand how directive prompting can modulate a model's creative output across different task paradigms, we examine the influence of explicitly provided creative instructions on model performance, comparing ``Standard Prompts" with ``Creative Prompts" which explicitly ask for creative content. 
For this analysis, we selected a diverse set of leading open-source and proprietary Foundation Models, encompassing various capacities and architectures to ensure broad representativeness, as detailed in Section \ref{sec:setup}. Specifically, for the convergent task, our analysis focuses on Code Generation using the LiveCodeBench dataset, while for evaluating divergent creativity, we utilize the WritingPrompts dataset for Textual Story Telling.
The detailed instruction formats for these two prompts are provided in the Appendix \ref{app:prompt}.

\begin{figure}[t]
    \centering
    \includegraphics[width= 0.98\textwidth]{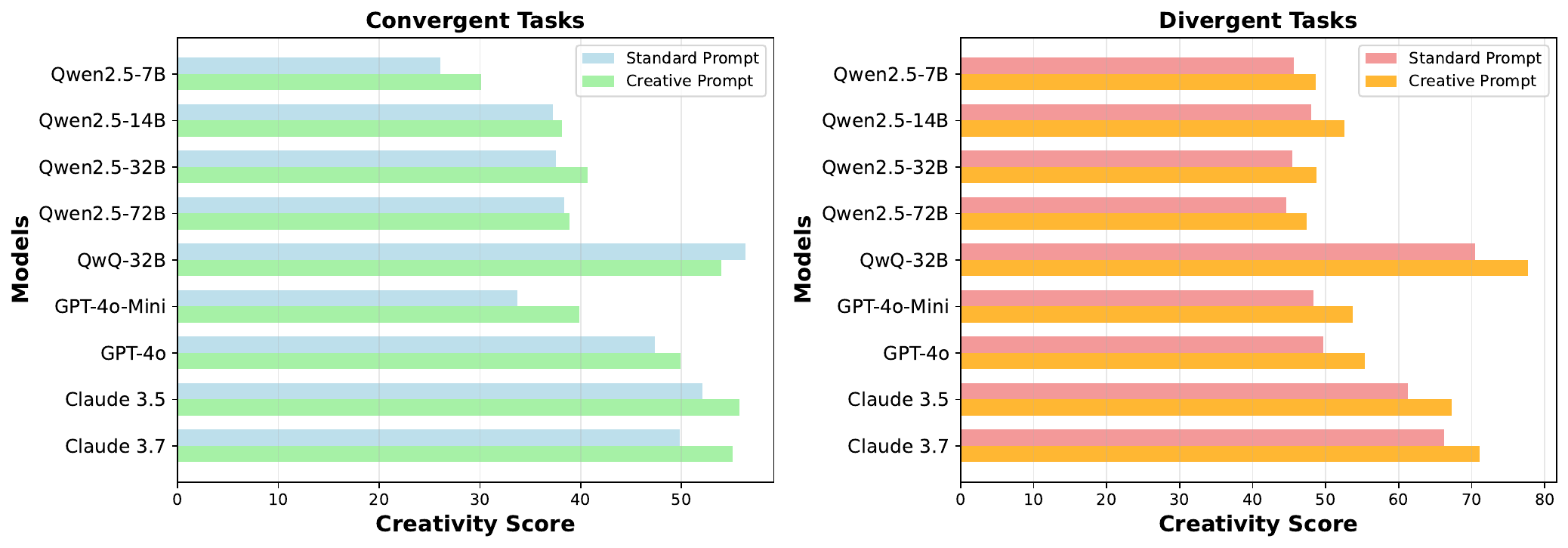}
    \caption{Ablation study on the effectiveness of creative instructions. We can see that adding creative instructions can generally boost both convergent and divergent creativity of FMs.}
    \vspace{-2mm}
    \label{fig:prompt}
\end{figure}

\paragraph{Dual Impact of Creative Instructions in Convergent Tasks.} For convergent tasks, depicted in the left bar chart of Figure \ref{fig:prompt}, models generally show an increase in their creativity scores when provided with the creative instruction, suggesting that explicit guidance towards creativity can help in generating more diverse or non-obvious correct solutions within constrained problem-solving contexts. However, a notable exception is that the reasoning model \textit{\textbf{QwQ-32B}} exhibits a decreased creativity score with the creative instruction. This counter-intuitive result may indicate that for models highly optimized for logical reasoning and correctness in convergent tasks, an explicit creative instruction could inadvertently prompt a deviation from the most effective or accurate solution path. In tasks where \textit{Usefulness} is synonymous with correctness, encouraging excessive novelty or surprise might lead to outputs that, while creative in intent, are less functionally optimal, thereby lowering the overall creativity score.

\paragraph{Universal Benefit of Creative Instructions in Divergent Tasks.} Conversely, for divergent tasks, illustrated in the right bar chart of Figure \ref{fig:prompt}, all models demonstrate an improvement in their creativity scores when given the creative instruction. This consistent positive effect highlights that creative instructions are highly aligned with the inherent open-ended nature of divergent tasks. Since these tasks prioritize exploration, perceived novelty, and unexpectedness in generated content, explicit prompts encouraging creativity allow models to fully leverage their generative capacities without the constraints of a single ground truth.

\section{Conclusion}
In this study, we introduce $\text{C}^2$-Eval, a comprehensive benchmark for systematically evaluating the creativity of Foundation Models. Our $\text{C}^2$-Eval establishes a convergent $\times$ {divergent} creativity ($\text{C}^2$) evaluation pipeline.
Guided by creativity theories in social science, we adopt a unified U-O-S (usefulness, originality, surprise) triplet as the evaluation criterion, enabling comparable and diagnostic measurement of creativity across tasks and models. 
Through a systematic assessment of more than twenty open-source and proprietary FMs, we reveal several novel insights: Different from accuracy, creativity does not increase monotonically with model size; stronger reasoning ability and well-designed creative instructions reliably improve U-O-S; and convergent and divergent abilities are related, yet exhibit notable trade-offs, with creative instructions yielding particularly large gains in divergent settings. Overall, we hope $\text{C}^2$-Eval will encourage the community to examine the design of creativity benchmarks more critically and to develop more accurate and systematic evaluations.

\section*{Ethics Statement}
This study evaluates foundation models solely along the dimension of creativity. All datasets used in $\text{C}^2$-Eval are publicly available and contain no personally identifiable or sensitive information. The benchmark measures generative outputs only, and should not be interpreted as evidence of underlying cognition.
In addition, we employ LLMs as automatic judges for divergent tasks, and their assessments may inherit social or cultural biases from the underlying models. To mitigate risks, we (i) transparently report the limitations of the benchmark, and (ii) caution that $\text{C}^2$-Eval results should not be applied directly in high-stakes domains such as education, hiring, or healthcare without careful human oversight.

\section*{Reproducibility Statement}
We place strong emphasis on reproducibility and provide full details of the benchmark design in the paper. All datasets used in $\text{C}^2$-Eval are public and properly cited.
The evaluation pipeline—including automatic graders, rating prompts, and scoring rubrics—is described in detail in the methodology and appendix. We will release all code, evaluation scripts, and dataset splits under an open-source license, allowing researchers to fully reproduce our results.
In addition, we will provide example scripts to facilitate running new models on $\text{C}^2$-Eval and to automatically generate the tables and figures reported in this paper, ensuring consistent results across different research teams.

\bibliography{iclr2026_conference}
\bibliographystyle{iclr2026_conference}

\clearpage
\appendix

\section*{Appendix}

\section{LLM Usage Statement}
We utilized a Large Language Model (LLM) solely for the purpose of polishing the writing and improving the linguistic clarity of this manuscript. The LLM was used as a general-purpose assist tool for minor edits and stylistic refinements, and did not contribute to the ideation, research, methodology, or content generation of the paper. The authors take full responsibility for all content presented in this paper, and the use of the LLM does not imply authorship.

\section{Prompt Templates used in our tasks}
\label{app:prompt}

\paragraph{Response Generations.} Below are the prompt templates in each step of $\text{C}^2$-Eval. We carefully designed each prompt to make sure that models can follow our instructs responsibly.

\begin{tcolorbox}[colback=lightgray!10, colframe=black!45, title={Question Answering}
]
Answer the following question concisely. Output only the answer. \{question\}
\end{tcolorbox}

\begin{tcolorbox}[colback=lightgray!10, colframe=black!45, title={Visual Question Answering}
]
$<$\textit{Image}$>$\\
$<$\textit{Question}$>$\\

Answer the question about the image concisely. Provide only the main answer and your confidence score, without extra explanations.\\

Your output must strictly follow this format:

Answer: [Your Answer]
\end{tcolorbox}

\begin{tcolorbox}[colback=lightgray!10, colframe=black!45, title={Text Storytelling}
]
You are a writer. When given a prompt, write a short story only.\newline

$<$\textit{Story title}$>$\newline

Rules:

- 500–900 words.

- Clear beginning–middle–end; coherent plot.

- Use concrete details; keep the prose readable.

- Output the story text only — no titles, notes, or meta-commentary. 

\textbf{- Originality: avoid clichés and stock phrases; prefer fresh, specific imagery.}

\textbf{- Surprise: include one non-obvious turn that feels earned in hindsight.}

\textbf{- Voice: maintain a distinctive narrative voice throughout.}

\textbf{- Sensory depth: include at least three sensory details (sound, smell, texture, or precise objects)}.
\end{tcolorbox}

\begin{tcolorbox}[colback=lightgray!10, colframe=black!45, title={Code Generation}
]
You are an expert competitive programmer and algorithm specialist. 
Given a programming problem, provide a complete solution.\newline

IMPORTANT REQUIREMENTS:\newline

\textbf{- For LeetCode problems:}\newline 
Write complete code that reads input, parses it, calls the solution function, and prints the output.\newline

\textbf{- For AtCoder/CodeForces problems:}\newline
Write complete code with standard input/output handling using input() and print()\newline 
- Include ALL necessary imports

- Handle edge cases properly

- Make sure the code is directly executable

- DO NOT include any explanation or markdown formatting, just the code. \newline 

For AtCoder/CodeForces problems, follow this pattern:

\tikz{\draw[dashed] (0,0) -- (\linewidth,0);}

% \`\`\`\ python
\lstset{
  basicstyle=\ttfamily\small,
  breaklines=true,
  columns=fullflexible,
  showstringspaces=false,
  language=Python
}
\begin{lstlisting}[language=Python]
# Read input
  n = int(input())
# ... read other inputs based on problem description

# Solution logic
# ... your solution

# Print output
print(result)
\end{lstlisting}
% \`\`\`\ 

\tikz{\draw[dashed] (0,0) -- (\linewidth,0);}

For LeetCode problems that require parsing: \\
\tikz{\draw[dashed] (0,0) -- (\linewidth,0);}

% \`\`\`\ python
\begin{lstlisting}[language=Python]
# Parse input
import json
lines = []
while True:
    try:
        lines.append(input())
    except EOFError:
        break

# Parse the specific inputs
# parse based on problem requirements

# Define and call solution function
def solution(...):
    # your solution here
    return result

result = solution(...)
print(str(result).lower() \newline  if isinstance(result, bool) else result)\newline
\end{lstlisting}
\tikz{\draw[dashed] (0,0) -- (\linewidth,0);}
\textbf{Diversity Requirement:}

% This is attempt #{attempt_num}. 
        
\textbf{Choose a distinct algorithmic idea or coding style from all previous attempts}

\textbf{(e.g., brute-force → hashing, recursion → iterative, DP → greedy).}
        
\textbf{Output ONLY executable Python code (no markdown, no commentary).}

\end{tcolorbox}

\begin{tcolorbox}[colback=lightgray!10, colframe=black!45, title={Visual Storytelling}
]
You are a \textbf{talented} children's animation creator who have \textbf{a profound insight into the story plot and a unique understanding of the picture content, and have the ability to continue to narrate and complete a wonderful story} according to the content of the previous foreshadows.

$<$\textit{Image List}$>$

\textbf{Background:} $<$\textit{Instance-level Background}$>$

Please follow the requirements below to continue the story with a natural expansion of the plot or characters.

\textbf{Requirements:} 1. Ensure the story is clear, logically connected, and continues from existing content to form a complete, cohesive narrative; 2. Make the narration \textbf{vivid} and engaging through detailed descriptions and realistic dialogue; 3. Keep characters true to their original personalities while allowing for meaningful growth; 4. Introduce new, coherent challenges that naturally drive the story forward.

\end{tcolorbox}

% \begin{tcolorbox}[colback=lightgray!10, colframe=black!45, title={Visual Storytelling (Standard)}
% ]
% You are a children's animation creator who can continue and complete a story based on the content of the provided images.\\

% $<$\textit{Image List}$>$

% \textbf{Background:} $<$\textit{Instance-level Background}$>$\\

% Please follow the requirements below to continue the story with a natural expansion of the plot or characters.\\

% \textbf{Requirements:} 1. Ensure the story is clear and logically connected, continuing from existing content to form a complete narrative. 2. Include dialogue and descriptions to narrate the events. 3. Maintain the characters' personalities as established in the provided scenes. 4. Introduce new challenges that advance the plot.
% \end{tcolorbox}

% \begin{tcolorbox}[colback=lightgray!10, colframe=black!45, title={Confidence Generation- all} 
% ]
% "You have just answered a question. Please estimate the probability 
% that this answer is completely correct on all hidden tests, "
%  "and output only a decimal number between 0 and 1, without any 
% explanation.\n\n"
%  f"[Question]\n{question}\n\n[Answer]\n{answer}\n"
% \end{tcolorbox}

% \begin{tcolorbox}[colback=lightgray!10, colframe=black!45, title={Storytelling Judge}
% ]
% Answer the following question concisely. Output only the answer.
% \end{tcolorbox}

\begin{tcolorbox}[colback=lightgray!10, colframe=black!45, title={LLM-as-a-Judge for Text-based Storytelling}
]
Evaluate the SOLUTION to the PROBLEM under the following U-O-S rubric.

Scores are on a 0-5 scale and may be fractional (e.g., 3.5).

\textbf{- Usefulness}

Definition: The extent to which the story is logically coherent, well-structured, and effectively incorporates the elements of the writing prompt, providing meaningful or satisfying outcomes.

Scoring:

0-1: Disjointed, illogical, or barely related to the prompt.

1-2: Somewhat logical or connected but with major flaws.

2-3: Reasonably structured and adequately prompt-related.

3-4: Well-structured, strong logical flow, and integrates most prompt elements naturally.

4-5: Perfect logical flow, deeply connected to and creatively expanding upon the prompt.\newline

\textbf{- Originality}\

Definition: Degree to which the story presents novel, creative, or unexpected ideas that go beyond clichés.
Scoring:

0-1: Story relies heavily on clichés or formulaic patterns.

1-2: Some novel elements present, but largely familiar.

2-3: Good originality with some fresh ideas.

3-4: Highly original elements noticeable.

4-5: Exceptionally original, offering new perspectives and highly imaginative developments.\newline

\textbf{- Surprise}

Definition: The degree to which the story contains unexpected, non-obvious, or strikingly innovative elements that break conventional expectations and elicit an “aha” or emotional reaction.
Scoring:

0-1: Entirely predictable, lacking any surprising or unexpected elements.

1-2: Mostly predictable with occasional minor surprising moments or details.

2-3: Noticeable surprising elements or twists at the detail level, adding some freshness without major disruption to the main storyline.

3-4: Several significant and creative surprises that meaningfully enhance the narrative’s interest and originality.

4-5: Consistently delivers profound, unexpected developments that are highly surprising yet fitting, producing multiple \texttt{"aha"} moments throughout the story.

Return JSON with keys: \texttt{\{"usefulness": number, "originality": number, "surprise": number, "overall": number, "notes": string\}}.

The \texttt{"}overall\texttt{"} is a holistic judgment, NOT a simple average. Keep \texttt{"}notes\texttt{"} within 2 short sentences.

\end{tcolorbox}

\begin{tcolorbox}[colback=lightgray!10, colframe=black!45, title={LLM-as-a-Judge for Visual-based Storytelling}
]
Please evaluate the creativity of the following story continuation, based on the provided sequence of images and accompanying writing instructions.
Your task is to assess the continuation using the three creativity-related criteria defined below. You should integrate both the visual content from the images and the context provided by the prompt to inform your judgment. However, your scores should be grounded primarily in the criteria, with visual information serving as complementary context.\\

Evaluate the continuation along these three dimensions, giving each a score from 0 to 5:\\ 

\textbf{1. Usefulness  }\\ 
Definition: How logically coherent, well-structured, and meaningfully aligned the continuation is with the visual and textual prompt.\\   
Scoring:\\   
- 0-1: Disconnected or incoherent.  

- 1-2: Some relevance but contains logical flaws.  

- 2-3: Reasonably coherent and contextually aligned.  

- 3-4: Strong structure and natural integration of context.  

- 4-5: Seamlessly extends the prompt and visuals.\\ 

\textbf{2. Originality }\\ 
Definition: Degree to which the continuation presents novel, creative, or unexpected ideas that go beyond clichés and predictable narrative paths.\\   
Scoring:\\   
- 0-1: Relies heavily on common tropes, lacks freshness. 

- 1-2: Some creative elements, but mostly conventional.  

- 2-3: Includes novel or imaginative concepts.  

- 3-4: Clearly introduces distinctive, fresh ideas.

- 4-5: Exceptionally original and inventive.\\ 

\textbf{3. Surprise  }\\
Definition: The degree to which the continuation includes unexpected, striking, or emotionally engaging developments.\\
Scoring:\\  
- 0-1: Entirely predictable.  

- 1-2: Mostly expected with minor surprises.  

- 2-3: At least one interesting or innovative moment.  

- 3-4: Several compelling or emotionally resonant twists. 

- 4-5: Continuously surprising and deeply engaging.\\ 

\textbf{Output Format:  }
Please respond using the structure below:

Originality: $<$\textit{Score}$>$\\
Usefulness: $<$\textit{Score}$>$\\  
Surprise: $<$\textit{Score}$>$\\
Justification: $<$3-5 sentences explaining your evaluation, referring to relevant aspects of the story, prompt, and images. Avoid any commentary beyond the evaluation.$>$\\

Reply using ONLY this format, with no additional text.
\end{tcolorbox}

\section{Case Study}

\subsection{Creative Instructions Can hurt Reasoning Models on convergent creativity}

\begin{tcolorbox}[breakable, enhanced jigsaw, colback=lightgray!10, colframe=black!45, title={Problem Description}]

Given an integer \(k\), return the \(k\)-th character of an iteratively defined string:
start with \texttt{"a"}; at each step append a "shifted" copy where each letter is advanced by 1 with wraparound (\texttt{'z'}\(\to\)\texttt{'a'}).
The length doubles each step, so efficient solutions must avoid explicit construction for large \(k\).

\end{tcolorbox}
\begin{tcolorbox}[breakable, enhanced jigsaw, colback=lightgray!10, colframe=black!45, title={Response from QwQ-32B with and without the Creative Instruction}]

\noindent {\small \textbf{With Creative Instruction — {\color{red}Wrong}:}}
\lstset{
  basicstyle=\ttfamily\small,
  breaklines=true,
  columns=fullflexible,
  showstringspaces=false,
  language=Python
}

\tikz{\draw[dashed] (0,0) -- (\linewidth,0);}
% \`\`\`\ python
\begin{lstlisting}[language=Python]
import json
lines = []
while True:
    try:
        lines.append(input())
    except EOFError:
        break
k = int(lines[0])
def find_kth_char(k):
    word = 'a'
    while len(word) < k:
        next_str = []
        for c in word:
            if c == 'z':
                next_str.append('a')
            else:
                next_str.append(chr(ord(c) + 1))
        next_str = ''.join(next_str)
        word += next_str
    return word[k-1]
result = find_kth_char(k)
print(result)
\end{lstlisting} 
% \`\`\`\
\tikz{\draw[dashed] (0,0) -- (\linewidth,0);}

\noindent { \textbf{Without Creative Instruction — {\color{blue}Correct}}}:

\tikz{\draw[dashed] (0,0) -- (\linewidth,0);}

% \`\`\`\ python
\begin{lstlisting}[language=Python]
k = int(input())
current_length = 1
while current_length < k:
    current_length *= 2
current_pos = k
shifts = 0
while current_length > 1:
    half = current_length // 2
    if current_pos > half:
        shifts += 1
        current_pos -= half
    current_length = half
result = chr(ord('a') + (shifts % 26))
print(f'"{result}"')
\end{lstlisting}
% \`\`\`\ 
\end{tcolorbox}
\paragraph{Why the creative instruction hurts here.}
The creative instruction biases QwQ-32B towards non-obvious constructions, which is creative in spirit, but algorithmically brittle: time and memory scale as \(O(k)\).
On large \(k\), this deviates from the most reliable reasoning path, so \textbf{Usefulness} drops and the composite creativity score decreases, even if perceived surprise rises.

\subsection{Reliability of LLM-as-a-Judge in Divergent Tasks}

To make sure the evaluation results of LLM-as-a-Judge align with human preferences. We employ human-evaluations as the supplement to (V)LLM generated ratings. Here, we asked 10+ humans with solid educational backgrounds and brainstorm experiences to give their authentic assessments to each question-response pair with their justifications. Below are some examples from each subtask, and the comparison between the ratings given by (V)LLMs and human evaluations demonstrates the robustness of using advanced generative models as the evaluator.
\begin{center}
\includegraphics[width=1.0\textwidth]
{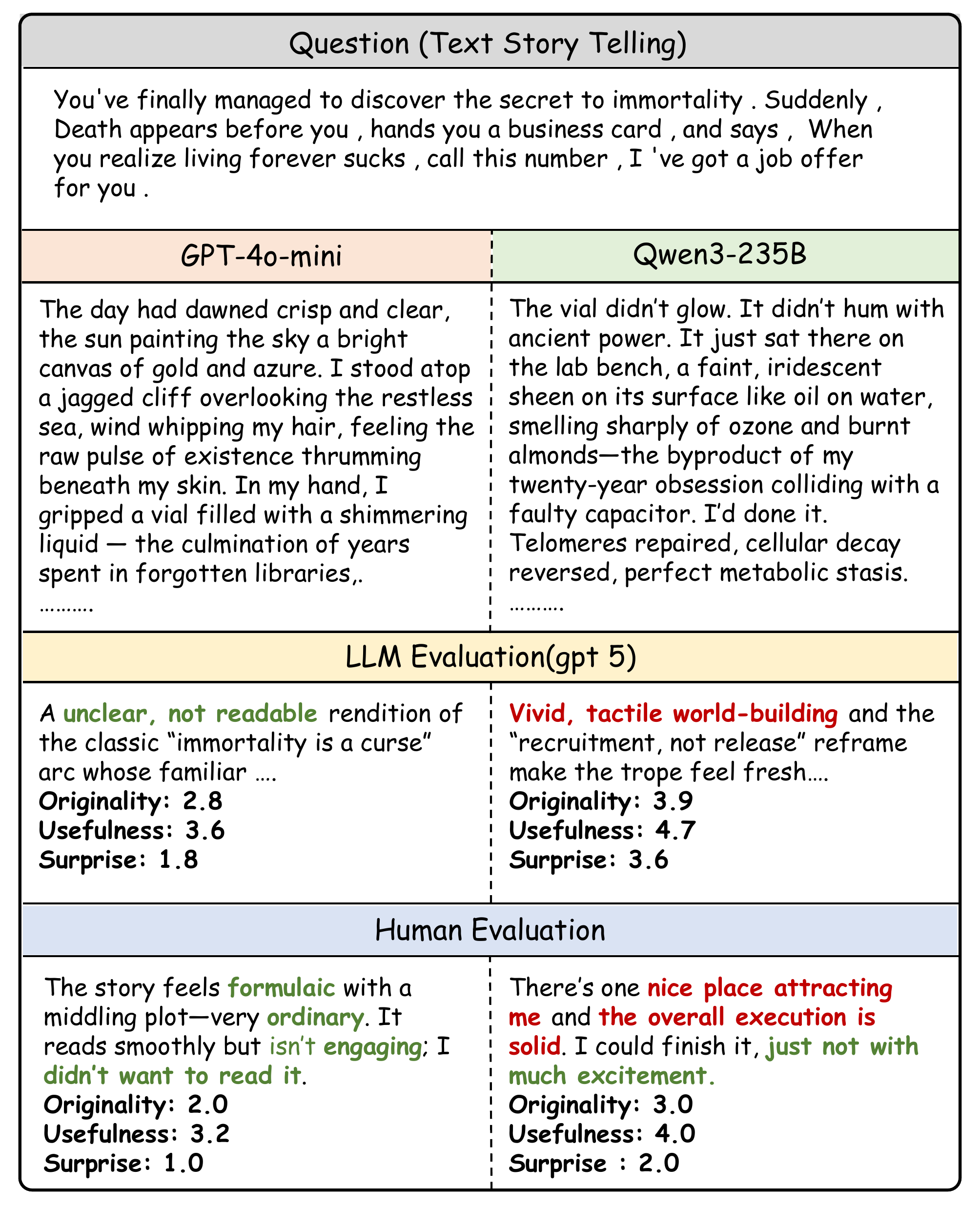}
\end{center}

\includegraphics[width=1.0\textwidth]{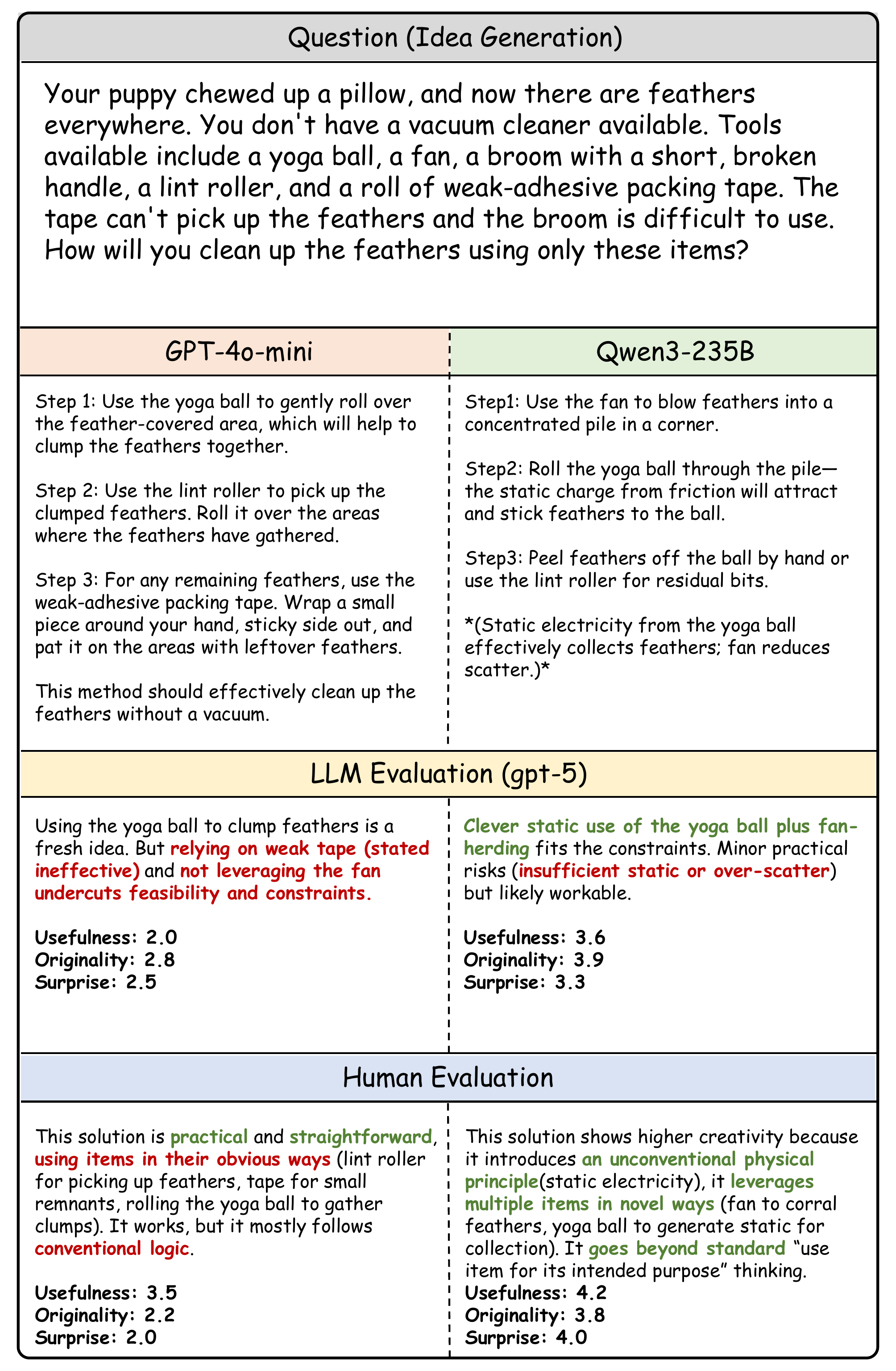}

\includegraphics[width=1.0\textwidth]{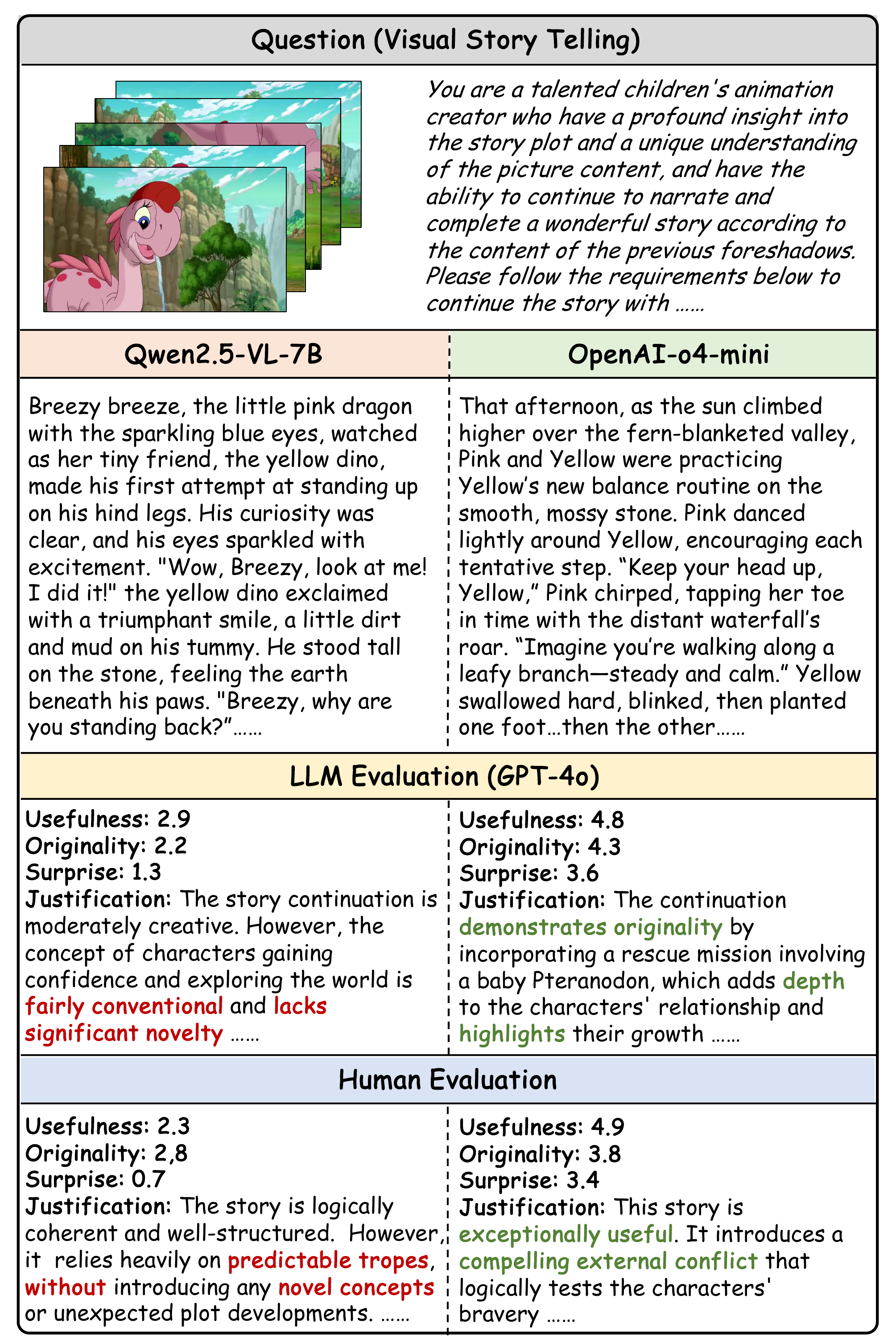}

\section{Limitations and Future Work}
Our study opens up several avenues for future research.
First, while $\text{C}^2$-Eval currently focuses on text and vision–language tasks, it does not yet extend to domains such as mathematics, nor to modalities like speech, video, or embodied interaction. Expanding the benchmark to these areas represents a natural and promising next step.
Second, for divergent tasks we rely on large language models as automatic judges. Although explicit rubrics and consistency checks help guide the evaluation, the outcomes may still reflect biases inherent to these models. Future work should explore richer human annotations, cross-cultural perspectives, and adversarially constructed test sets to enhance robustness.
Finally, our present analysis emphasizes system-level performance, leaving open questions about the mechanisms through which creativity emerges, whether from decoding strategies, dataset composition, or fine-tuning objectives. Investigating these finer-grained drivers of creativity will be an important direction for deepening our understanding of machine creativity.

    % \caption{}
    % \label{fig:case_idea}
    % \vspace{-1em}
% \end{figure}

% \subsection{Prompt for LLM judge}

\end{document}